# Prediction of the outcome of a Twenty-20 Cricket Match

Arjun Singhvi, Ashish V Shenoy, Shruthi Racha, Srinivas Tunuguntla

*Abstract*—Twenty20 cricket, sometimes written Twenty-20, and often abbreviated to T20, is a short form of cricket. In a Twenty20 game the two teams of 11 players have a single innings each, which is restricted to a maximum of 20 overs. This version of cricket is especially unpredictable and is one of the reasons it has gained popularity over recent times. However, in this project we try four different approaches for predicting the results of T20 Cricket Matches. Specifically we take in to account: previous performance statistics of the players involved in the competing teams, ratings of players obtained from reputed cricket statistics websites, clustering the players' with similar performance statistics and using an ELO based approach to rate players. We compare the performances of each of these approaches by using logistic regression, support vector machines, bayes network, decision tree, random forest and their

*Keywords—clustering; ELO[8];*

## I. INTRODUCTION

A typical Twenty20 game is completed in about three hours, with each innings lasting around 75–90 minutes and a 10–20-minute interval. Each innings is played over 20 overs and each team has 11 players. This is much shorter than previously existing forms of the game, and is closer to the timespan of other popular team sports. It was introduced to create a fast-paced form of the game, which would be attractive to spectators at the ground and viewers on television.

Since its inception the game has been very successful resulting in its spread around the cricket world and spawned many premier cricket league competitions such as the Indian Premier League. On most international tours there is at least one Twenty20 match and all Test-playing nations have a domestic cup competition.

One of the International Cricket Council's (ICC) main objectives in recent times is to deliver real-time, interesting, storytelling stats to fans through the Cricket World Cup app or website. Players are what fans obsess most about so churning out information on each player's performance is a big priority for ICC and also for the channels broadcasting the matches.

Hence to solving an exciting problem such as determining the features of players and teams that determine the outcome of a T20 match would have considerable impact in the way cricket analytics is done today. The following are some of the terminologies used in cricket: This template was designed for two affiliations.

*1) Over:* In the sport of cricket, an over is a set of six balls bowled from one end of a cricket pitch.

*2) Average Runs:* In cricket, a player's batting average is the total number of runs they have scored, divided by the number of times they have been out. Since the number of runs a player scores and how often they get out are primarily measures of their own playing ability, and largely independent of their team mates, batting average is a good metric for an individual player's skill as a batsman.

*3) Strike Rate:* The average number of runs scored per 100 balls faced. (SR = [100 * Runs]/BF).

*4) Not Outs:* The number of times the batsman was not out at the conclusion of an innings they batted in.

*5) Economy:* The average number of runs conceded per over.(Economy=Runs/Overs bowled).

*6) Maiden Over*: A maiden over is one in which no runs are scored.

*7) No Balls:* In domestic 40-over cricket, a no-ball concedes two runs. In Twenty20 cricket, a no-ball is followed by a 'free hit', a delivery from which the batsman can not be bowled or caught out, but can still be run out.

*8) Wides*: In the sport of cricket, a wide is one of two things:
  a) The event of a ball being delivered by a bowler too wide or (in international cricket) high to be hit by the batsman, and ruled so by the umpire.
  b) The run scored by the batting team as a penalty to the bowling team when this.

## II. DATASET PREPARATION

For the problem being solved at hand, since there were no datasets readily available we had to identify the right sources to collect data from, followed by crawling the source web pages to collect raw data extract required information from the raw data and finally construct features from this data.

### A. Source

The data used to generate the player statistics in this project was obtained from espncricinfo.com, which provided us the most comprehensive cricket coverage available including, live ball-by-ball commentary. In this project the data of only international, domestic and league T20 cricket matches was used since this would be the closest representation of how a player would perform in a T20. A list of 5390 T20 matches was obtained from ESPN. The dataset for all the classifiers was this set of 5390 matches.

### B. Crawling

The next task was to obtain the scorecard of each of these matches along with statistics such as the number of extras, players who did not bat, location and the date when the match was played. To do this we built a web crawler using python and BeautifulSoup library to extract and parse the data from the match URLs to our required format.

### C. Feature Extraction

To decide the features needed to train the machine learning algorithms, we decided to survey the parameters Cricket experts use to arrive at player ratings in espncricinfo.com, cricmetric.com. Along with the features that the experts usually use we also added features such as "CurrentBattingPosition" and "AverageBattingPosition".

To make sure that only the statistics before the date when the match was played is taken into consideration to determine the averages we designed python functions to be called on the crawled cricinfo data. The functions would take in a date parameter and determine the player's statistics from matches played on or before that particular date.

The final set of 16 features per player that we decided to use in this project is as shown in Table I. This set of features is applicable to each player in the team and is a combination of both bowling and batting parameters. A subset or the entire set of these features was used in each approach we followed to tackle the problem.

| Feature Name | Feature Description |
|---|---|
| Average Runs | Continuous |
| Average Number of 4s | Continuous |
| Average Number of 6s | Continuous |
| Average Strike Rate | Continuous |
| Number of Not Outs | Continuous |
| Number of 50s | Continuous |
| Number of 100s | Continuous |
| Number of Matches | Continuous |
| Current Batting Position | Continuous |
| Average Batting Position | Continuous |
| Average Number of Wickets taken per Match | Continuous |
| Average Economy | Continuous |
| Average Runs Conceded | Continuous |
| Average Number of Wides | Continuous |
| Average Number of Balls | Continuous |
| Average Number of Maiden Overs | Continuous |
| Classification Value | 0/1 for team 1 win/ team 2 win respectively |

TABLE 1 : LISTS THE DATA FEATURES AND THEIR DESCRIPTION

## III. CLASSIFIERS

### A. AdaBoost

An *AdaBoost classifier*[1] is a meta-estimator that begins by fitting a classifier on the original dataset and then fits additional copies of the classifier on the same dataset but where the weights of incorrectly classified instances are adjusted such that subsequent classifiers focus more on difficult cases. We used this class from scikit-learn machine learning python package[2] that implements the algorithm known as AdaBoost-SAMME [3].

### B. Decision Trees

Decision Trees (DTs) are a non-parametric supervised learning method used for classification and regression. The goal is to create a model that predicts the value of a target variable by learning simple decision rules inferred from the data features. In decision tree, leaves represent class labels, internal nodes represent features and the branches represent the value of the feature in the internal nodes. Scikit-learn[2] package provides a *DecisionTreeClassifier* which is the implementation for a decision tree.

### C. Naïve Bayes

Naive Bayes methods are a set of supervised learning algorithms based on applying Bayes' theorem with the "naive" assumption of independence between every pair of features. In spite of their apparently over-simplified assumptions, naive Bayes classifiers have worked quite well in many real-world situations[4]. They require a small amount of training data to estimate the necessary parameters.

Naive Bayes learners and classifiers can be extremely fast compared to more sophisticated methods. The decoupling of the class conditional feature distributions means that each distribution can be independently estimated as a one dimensional distribution. This in turn helps to alleviate problems stemming from the curse of dimensionality. We used the implementation provided by Scikit-learn [2] for this.

### D. Random Forests

Random Forests adopt the ensemble learning method where many weak learners are combined to produce a strong learner. It is a meta estimator that fits a number of decision tree classifiers on various sub-samples of the dataset and use averaging to improve the predictive accuracy and control over-fitting[5]. The sub-sample size is always the same as the original input sample. We used the implementation provided by Scikit-learn [2] for this.

### E. Support Vector Machines

Support vector machines are a set of supervised learning methods used for classification, regression and outliers. Support Vector Machine is a non-probabilistic binary classifier. It learns a maximum margin separating hyperplane for classification. For data that is not linearly separable, there are two approaches to learn the hyperplane, linear SVM with slack variables or kernel based SVM. For linear SVM it introduces slack variables to learn a separating plane with relaxation whereas for kernel based SVM data is transformed into a different space and hyperplane is constructed in the new space for classification. For our data set we tried both these approaches(Linear and Non-linear) using the implementation provided by Scikit-learn[2].

### F. Bagging

Bagging predictors is a method for generating multiple versions of a predictor and using these to get an aggregated predictor. The aggregation averages over the versions when predicting a numerical outcome and does a plurality vote when predicting a class. The multiple versions are formed by making bootstrap replicates of the learning set and using these as new learning sets[6]. Tests on real and simulated data sets using classification and regression trees and subset selection in linear regression show that bagging can give substantial gains in accuracy. The vital element is the instability of the prediction method. If perturbing the learning set can cause significant changes in the predictor constructed, then bagging can improve accuracy. We used the implementation of Bagging provided by Scikit-learn [2] in combination with each of the above classifiers mentioned.

## IV. MULTIPLE APPROACHES FOR THE PREDICTION PROBLEM

### A. Approach 1

The feature values obtained as per Section II.C were normalized to obtain a value between 0 and 1 for each of the features. So each training set instance had a total of 16 features per player*22 players that totaled to 352 feature values. For debutant players or a player without a statistic for a particular feature, a normalized base rating of 0.5 was assigned.

*1) Variations*

Overall there were four variations based on varying training set size used to train the classifiers on the player performance statistics data set.

*a) Variation 1:* Ran the classifiers on the entire 352 feature values per training instance.
*b) Variation 2:* Aggregate each player's normalized statistics to arrive at two values - Batting Aggregate and Bowling Aggregate. In this approach each training instance would have 44 features.
*c) Variation 3:* Aggregate each player's batting aggregate and bowling aggregate to arrive at one value per player. In this approach each training instance would have 22 features.
*d) Variation 4:* Aggregate each team's normalized player statistics to arrive at one value per team. In this approach each training instance would just have 2 values. Team 1 aggregate and Team 2 aggregate.
*e) Variation 5:* To reduce the number of features in our training data set, we decided to use recursive greedy backward elimination. Before performing the backward elimination we decided to remove certain parameters that had mostly missing data or same values for most of the instances. These features were:

*NoOfNotOuts, AvgRunsConceded, AvgNoOfWides, AvgNoOfNoBalls*

We also decided to combine the *AverageBattingPos* and *CurrentBattingPos* to a single negative value if the difference between the two features was greater than a threshold of 2.

To do the backward elimination we used Adaboost and Randomized Forest classifiers as these algorithms had a consistently good performance on the initial set of features in the previous four approaches. The best accuracy was provided by randomized forest and the results obtained are as shown in the figure 1.

The features that were eliminated in this process before hitting a high of 59.0103% accuracy were:
1. Current Batting Position
2. Average Batting Position
3. NoOfMatches
4. NoOf100s

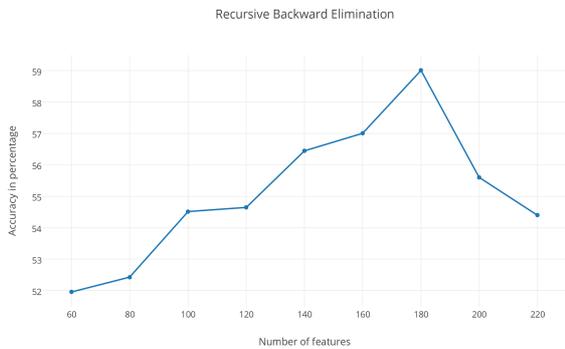

Fig. 1. Percentage Accuracy versus number of features during Backward Elimination

*2) Evaluation*

To evaluate the performance of different classifiers on the performance statistics feature set, we used a 10 fold Cross Validation. The data set obtained from 5390 matches was brought down to 796 to include only Indian Premier League T20 matches, to keep instances with missing statistics to a minimum.

The accuracies obtained after using the initial set of 378 features per training instance is as shown in table 2. The maximum accuracy obtained for predicting the outcome of a match in this variation was 53.4% after using a Decision Tree classifier with Bagging.

In the second variation where 44 features were used per training instance, we obtained an accuracy of 56.63% with an Adaboost classifier. This was an improvement from the previous approach and hence we decided to aggregate the features to 22 per training instance.

In this third variation the accuracy was around 54%, which was not much of an improvement from the original approach of using 378 features. Using the fourth variation of using 2 aggregated features per training instance did not improve the accuracy from the initial feature set as well.

| Classifiers | Variation 1 | Variation 2 | Variation 3 | Variation 4 |
| --- | --- | --- | --- | --- |
| Adaboost | 52.76 | 56.63 | 52.52 | 51.5 |
| Randomized Forest | 51.77 | 55.4 | 51.4 | 51.5 |
| Multinomial Naive Bayes | 51.15 | 51.14 | 51.76 | 50.75 |
| Multinomial Naïve Bayes(Bagging) | 51.65 | 51.6 | 52.15 | 51 |
| Decision Trees | 51.87 | 53.9 | 52.13 | 51.3 |
| Decision Trees(Bagging) | 53.4 | 52.65 | 52.4 | 53.41 |
| Linear SVM | 50.87 | 53.16 | 52.65 | 51.76 |
| Linear SVM(Bagging) | 50.24 | 50.25 | 51.65 | 53.52 |
| Non-Linear SVM | 50.59 | 50.25 | 52 | 52.39 |
| Non-Linear SVM (Bagging) | 50.4 | 53.01 | 54.14 | 50.24 |

TABLE 2 : LISTS THE ACCURACIES FOR THE DIFFERENT VARIATIONS FOR THE CANDIDATE CLASSISIFERS

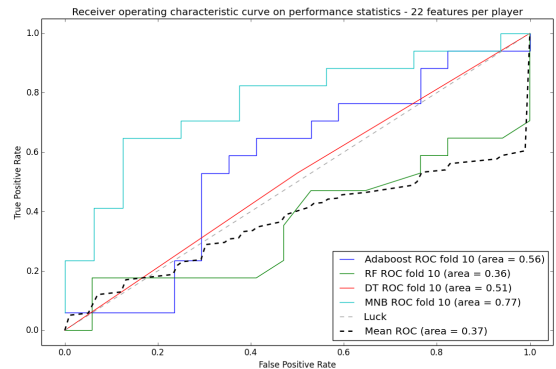

Fig. 2. ROC curve for Variation 3

As seen from the Figure 2, Naïve Bayes outperforms the other candidate classifiers

B. *Approach 2*

*1) Modelling player batting and bowling ratings:*

Most of the existing methods for ranking Cricket players are based on the aggregated statistics of players: total runs/total wickets, batting/bowling average, strike rate/economy rate etc. All of these methods suffer from one big flaw, which is that any player can have inflated statistics by playing against "weaker" teams, and therefore achieve a higher rating which may not reflect the true standing of the player. In this approach we present a new rating method which automatically takes care of the "strong" and the "weak" opponents, and is therefore more reflective of the true rating of the player in relation to his peers.

We consider the pairwise interactions between batsmen and bowlers, convert the pairwise interactions into a "score" and use these to rank the players. A score for an interaction between batsman *i* and bowler *j* is modelled as:
$$s_{ij} = A + a_i - b_j + \epsilon$$
where $a_i$ is the batting rating of player *i* and $b_j$ is the bowling rating of player *j*. The intercept A represents the expected score between average players. $\epsilon$ is a zero-mean random error. The error term is included because two competing players are not necessarily expected to repeat the same score in two different interactions.

The ratings represent the relative strength of each player. Suppose, player *i* has a batting rating of 10. It suggests that he would score 10 runs more than an average batsman. Similarly if a player has bowling rating of -10, it suggests that he would give 10 runs more than an average bowler.

The scores $s_{ij}$ are obtained using the *runs above average (RAA)* statistic. It has two parts, one which captures the strike rate of the batsman and the second which captures the rate at which a batsman gets out. The first part is computed for batsmen as follows. In 2011, the average batsman scored 0.79 runs per ball. If a batsman scores 900 runs in 1000 balls, he is scoring 900–0.79*1000 runs above average, which is 110.

The second part is computed by looking at the rate at which the batsman gets out. In 2011, the average batsman made 0.028 outs per ball. If the particular batsman we discussed above, in addition, made 20 outs in 1000 balls, then he made 0.028*1000–20 outs less than the average player over 1000 balls, which is 8. The overall average of ODI batsmen in 2011 was 28.31. So overall, he contributed 8*28.31=226.5 runs more than the average player, in terms of getting out. Taking the sum of both the parts, this particular batsman made 110+226.5=336.5 runs above average.

*2) Time Scaling*

While calculating the ratings of players, old performances are weighted less than the recent ones. The reason is that the performance of any player changes over time and capturing this temporal dependence is important for any rating system.
Let $t_{min}$ be the time of the earliest match in the dataset and $t_{max}$ be the time of the latest match.

For a game that happened at time *t*, the following weight worked the best.
$$w = \left(\frac{1 + t - t_{min}}{1 + t_{max} - t_{min}}\right)^2$$
During the training of ratings, each match's importance is scaled with the corresponding weight. This idea is inspired by Chessmetrics [7].

*3) Neighbours*

To avoid overfitting on the data, we restricted the dataset to contain only International matches and matches from Indian Premiere League, where the ratings of players are highly correlated to the opponents'. Let's define the neighborhood $N_i$ of a player *i* as the multiset of all opponents (opponent bowlers if player *i* is a batsman and vice versa) player *i* has played against. It's defined as a multiset, since a player may have played the same opponent many times. For every player *i*, with a neighborhood $N_i$ we use the weighted average $n_i$ of the ratings for each player in $N_i$, where the weights are calculated as described in the section "Time Scaling".

*4) Calculation of Ratings*

The cost function between the predicted and actual scores is defined as
$$cost = \sum_{i,j} w_{ij}(o'_{ij} - o_{ij})^2 + \lambda \sum_i (r_i - n_i)^2$$
where $w_{ij}$ is the weight of the match in which the interaction between players i ,j took place, $o'_{ij}, o_{ij}$ represent the predicted and actual scores for these interactions. The second term in the cost function ensures that the ratings of each player do not diverge a lot from their corresponding $n_i$ unless there is enough statistical evidence. The regularization parameter that worked best (found using cross validation) was 0.7.

The cost function takes into account the most recent of the interaction, the difference between the actual and predicted of outcomes and the difference between rating of the player and his neighbors. We used stochastic gradient descent approach to find the optimal ratings for players that minimize the cost function.

*5) Quality Comparison*

In this section we compare the ratings obtained against the ratings given by International Cricket Council (ICC). Note that ICC ratings are not available for all players.

First, we normalize the ratings obtained from the model described above to be on the same scale as the ICC ratings. The normalization helps compare directly the ratings.

Figure 3 and 4 are the scatter plots of ICC ratings and ratings we derived through approach 2. We observe that the distribution of ratings is similar and there is a very strong correlation for high ratings. In other words, players with high ICC ratings have also high ratings in our model and vice-versa.

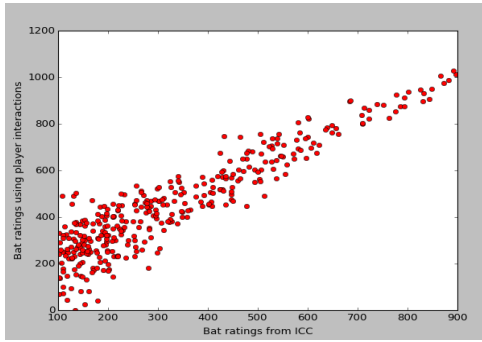

Fig. 3. Scatter Plot of batting ratings found by approach 2 vs ICC ratings

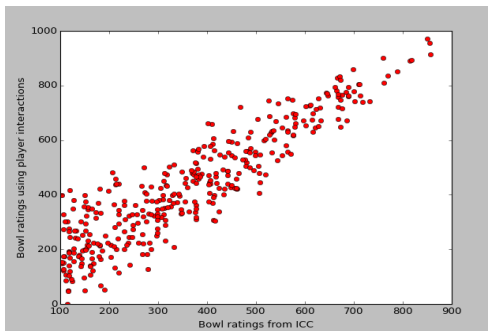

Fig. 4. Scatter Plot of batting ratings found by approach 2 vs ICC ratings

*6) Evaluation*

As seen from table 3 the ensemble of all the candidate classifiers performs the best. While comparing the individual classifier performances with the ratings derived in this approach, Non Linear SVMs tend to outperform the remaining classifiers in terms of mean accuracy measures as a performance metirc.

| Classifiers | Mean Accuracy(%) |
| --- | --- |
| Adaboost | 57.74 |
| Randomized Forest | 62.70 |
| Multinomial Naive Bayes | 57.23 |
| Multinomial Naïve Bayes(Bagging) | 58.34 |
| Decision Trees | 60.89 |
| Decision Trees(Bagging) | 60.36 |
| Linear SVM | 60.84 |
| Linear SVM(Bagging) | 61.77 |
| Non Linear SVM | 63.05 |
| Non Linear SVM(Bagging) | 63.89 |
| Ensemble of all the above | 64.62 |

TABLE 3 : LISTS THE ACCURACIES FOR THE DIFFERENT CARIATIONS FOR THE CANDIDATE CLASSISIFERS

## C. Approach 3 – Clustering to derive player ratings

### 1) Methodology

In this approach, the problem of match outcome prediction is solved using two stages as seen in Figure 5. The first stage involves coming up with two ratings (batting rating, bowling rating) per player based on the features mentioned in Table 1. Two separate k-means clustering systems are used to generate the batting and bowling ratings for each player. The k-means clustering system responsible for generating batting rating is trained using all the batting features for each player whereas the other system is trained by representing each player via all the bowling features. Using such a mechanism (to generate ratings) which involves taking into account a wide variety of features to represent players is a more correct representation of the players, rather than the available player ratings that are calculated using a static formula. The generation of ratings per player is a one-time step. At the end of this stage, we will be able to represent each player via the two generated ratings.

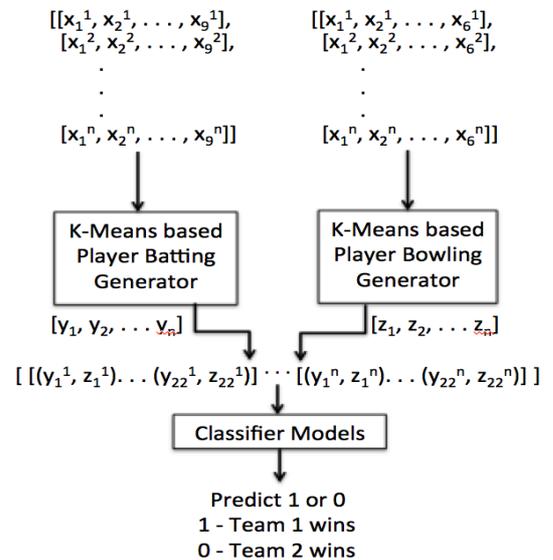

Fig. 5. Schematic diagram of approach 3

The second stage of the system does the actual task of match outcome prediction. This stage consists of a simple binary classifier as the prediction task eventually boils down to a classification. The instances to the classifier are of similar form as used in the above two models. The only difference being that instead of using ratings generated from a statistical formula, we would be using the ratings generated from the k-means clustering systems.

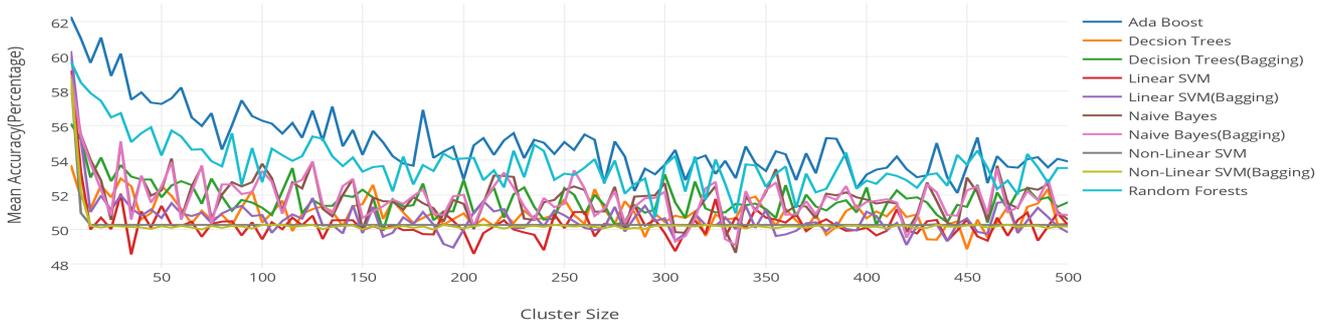

Fig. 6. Mean Accuracy vs Cluster Size for all candidate classifiers

This entire system consists of various parameters, which can be varied. In the first stage, choosing the number of clusters (that effectively translates to the range of valid rating) is one such parameter. The accuracy of the system is also affected based on the classifier used in the second stage. We have carried out experiments to choose the appropriate number of clusters for the various candidate classifiers chosen in the second stage. The results of these experiments are described in the next section.

*2) Evaluation*

Having decided on the candidate classifier as mentioned in section IV, we went ahead to use 10-fold internal cross-validation to choose the appropriate number of clusters for the 1st stage based on the accuracy of the second stage. As seen from Figure 8, the accuracy of all the candidate classifier is the best when the number of clusters is set to 5 and decreases thereafter. The reason for this is probably due to the fact that there isn't adequate data for representing each cluster.

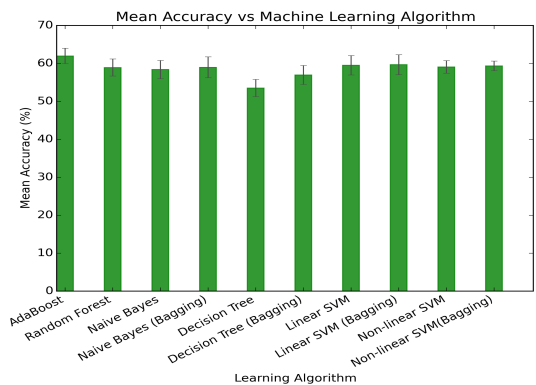

Fig. 7. Percentage Accuracy versus Classifier

Post fixing on the number of clusters in the first stage, we went ahead and evaluated the performance of the second stage classifier. As seen from Figure 7., AdaBoost gives the best accuracy of 62% whereas Decision Trees give the least accuracy of around 52%. Another trend observed is that Bagging doesn't increase the accuracy nor does it have a negative impact on most of the classifiers. This behavior can be attributed to the fact that most of the candidate classifiers are relatively stable, with Decision Trees being an exception. In Decision Trees, Bagging does have a noticeable positive impact, as the Decision Trees are unstable learning methods.

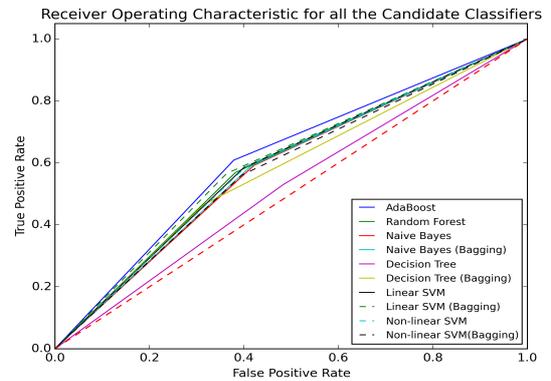

Fig. 8. ROC curve for all the classifiers

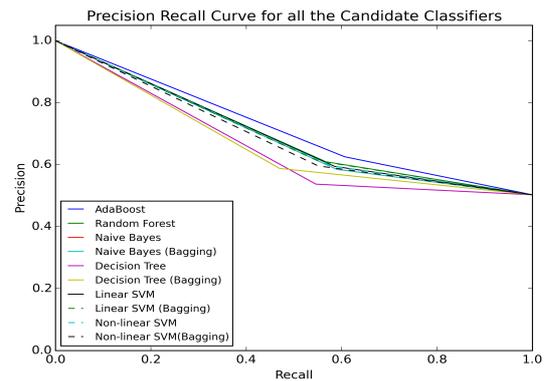

Fig. 9. Precision Recall Curver for all the Classifiers

The ROC curve corresponding to the various classifiers, as seen in Figure 8., show that all the candidate classifiers perform better than a system that does random guessing. Also, as expected, AdaBoost has the maximum AUC, which implies that it is the most efficient classifier in the second stage. Similarly, the Precision-Recall curve indicates that AdaBoost classifier outperforms all the other considered classifiers.

However, as seen from Figure 9., there is still room for improvement as the ideal point in the

Precision-Recall curve would in the upper right-hand corner.

## V. LIMITATIONS AND FUTURE WORK

The main limitation in carrying out this project was the limited dataset, which we had at our disposal. The next logical step in the direction to improve the accuracy of prediction problem at hand would be to test out the approaches and various methodologies proposed in this paper using a larger and more representative dataset. Also we would like to extend the candidate classifier set considered to a more exhaustive list and compare the performances among them.

## ACKNOWLEDGMENT

We would like to thank Prof. Mark Craven for his constant guidance during the course of the project.